\documentclass[letterpaper]{article} 
\usepackage{aaai2026}  
\usepackage{times}  
\usepackage{helvet}  
\usepackage{courier}  
\usepackage[hyphens]{url}  
\usepackage{graphicx} 
\usepackage{natbib}  
\usepackage{caption} 
\usepackage{booktabs}
\usepackage{amsmath}
\usepackage{amssymb}  
\usepackage{algorithm}
\usepackage{algorithmic}
\usepackage{colortbl}
\usepackage{newfloat}
\usepackage{listings}
\DeclareCaptionStyle{ruled}{labelfont=normalfont,labelsep=colon,strut=off} 
\floatstyle{ruled}
\newfloat{listing}{tb}{lst}{}
\floatname{listing}{Listing}
\nocopyright

\title{LogicPuzzleRL: Cultivating Robust Mathematical Reasoning in LLMs via Reinforcement Learning}
\author{
    Zhen Hao Wong\equalcontrib,
    Jingwen Deng\equalcontrib,
    Runming He,
    Zirong Chen,
    Qijie You, \\
    Hejun Dong,
    Hao Liang,
    Chengyu Shen,
    Bin Cui,
    Wentao Zhang
}
\affiliations{
    \textsuperscript{\rm 1}Peking University\\

    zhenhao1141@stu.pku.edu.cn
    wentao.zhang@pku.edu.cn
}

\usepackage{bibentry}

\begin{document}

\maketitle

\begin{abstract}
Large language models (LLMs) excel at many supervised tasks but often struggle with structured reasoning in unfamiliar settings. This discrepancy suggests that standard fine-tuning pipelines may instill narrow, domain-specific heuristics rather than fostering general-purpose thinking strategies. In this work, we propose a “play to learn” framework that fine-tunes LLMs through reinforcement learning on a suite of seven custom logic puzzles, each designed to cultivate distinct reasoning skills such as constraint propagation, spatial consistency, and symbolic deduction. Using a reinforcement learning setup with verifiable rewards, models receive binary feedback based on puzzle correctness, encouraging iterative, hypothesis-driven problem solving. We demonstrate that this training approach significantly improves out-of-distribution performance on a range of mathematical benchmarks, especially for mid-difficulty problems that require multi-step reasoning. Analyses across problem categories and difficulty levels reveal that puzzle training promotes transferable reasoning routines, strengthening algebraic manipulation, geometric inference, and combinatorial logic, while offering limited gains on rote or highly specialized tasks. These findings show that reinforcement learning over logic puzzles reshapes the internal reasoning of LLMs, enabling more robust and compositional generalization without relying on task-specific symbolic tools.
\end{abstract}

\begin{links}
    \link{Code}{https://github.com/wongzhenhao/GameRL}
\end{links}

\section{Introduction}
Large language models (LLMs) have made remarkable strides in recent years, particularly following extensive supervised fine-tuning (SFT) and reinforcement learning (RL)~\cite{guo2025deepseek,peng2023instruction,team2025kimi,xu2025towards}. These models, exemplified by GPT-4 and related architectures, demonstrate strong performance on in-distribution (ID) tasks such as code generation and standardized mathematics benchmarks, and are often described as possessing emergent reasoning abilities. However, despite their success on complex tasks, these same models frequently underperform on seemingly simple puzzle games that require structured logical thinking, spatial manipulation, or abstract pattern recognition~\cite{lin2025zebralogic}. These are skills that humans readily apply even in unfamiliar settings. This inconsistency suggests that existing SFT and RL regimes primarily impart narrow, domain-specific heuristics rather than fostering truly generalizable reasoning strategies capable of transferring to out-of-distribution (OOD) tasks.

In contrast, human learners develop a toolkit of abstract thinking strategies such as hypothesis testing, stepwise reasoning, and constraint satisfaction. These strategies enable them to tackle a broad spectrum of problems. For instance, the logical heuristics used to solve a sliding-tile puzzle can later support deductive reasoning in geometry or combinatorics. From this perspective, the critical question is not merely whether an LLM can perform well on a given benchmark, but how it arrives at its solutions, and whether it develops an internal reasoning process that applies across disparate domains.

In this work, we introduce a “play-to-learn” paradigm designed to shape the reasoning mechanisms of LLMs through engagement with structured puzzles. We create a suite of seven custom puzzle games, each with a unique, verifiable solution and each targeting a distinct reasoning archetype, including combinatorial logic, spatial manipulation, sequence deduction, and pattern recognition. These puzzles are deliberately constructed so that no background knowledge or memorized templates suffice. Instead, the model must reason through each step to reach the correct answer. Within a reinforcement-learning-with-verifiable-reward (RLVR) framework, the LLM iteratively generates candidate solutions and receives binary feedback (correct or incorrect), which provides a clear and interpretable signal for improvement. By using verifiable rewards, we encourage the model to adopt an iterative reasoning process involving hypothesis generation, intermediate checking, and corrective revision. This approach reflects the cognitive strategies that humans employ when acquiring new problem-solving skills.

Our central objective is not only to achieve quantitative gains on downstream tasks, but also to examine the qualitative transformation of the model’s internal reasoning. To this end, we conduct a comprehensive evaluation across a range of mathematical benchmarks, covering diverse categories such as arithmetic, algebra, and combinatorics, and spanning multiple difficulty levels from elementary problems to olympiad-style challenges. We find that LLMs trained with our seven-puzzle curriculum achieve statistically significant improvements across most categories and levels. Furthermore, through a detailed analysis of intermediate reasoning traces, including metrics such as the length and structure of generated solutions, the frequency of self-corrections, and the propagation of errors, we show that these gains result from improved reasoning ability rather than the accumulation of new domain-specific heuristics. In other words, puzzle-based training helps develop a more robust and systematic problem-solving framework, enabling the model to address novel mathematical problems with greater precision and fewer random guesses.

This paper first situates our work within the broader context of efforts to improve LLM reasoning, reviewing recent progress in RLVR, tool-augmented reasoning, and synthetic puzzle frameworks. We then describe the design principles of our seven puzzle games, including their verifiable reward functions, incremental curricula, and targeted reasoning skills. Next, we outline our RL training protocol, detailing the choice of base LLM, the reward shaping strategies, and the exploration incentives such as entropy-based losses. We present empirical results evaluating (1) in-domain puzzle performance, (2) out-of-domain mathematical problem-solving ability, and (3) ablation studies to isolate the contributions of individual puzzle archetypes. Finally, we perform a qualitative analysis of the model’s evolving reasoning behavior by examining the frequency of reflection, patterns in error correction, and strategies used in hypothesis generation. These insights help clarify how puzzle-based RL fosters robust, transferable thinking skills.

Through this “play-to-learn” framework, we show that engaging LLMs with small, verifiable puzzles can restructure their internal reasoning processes and improve performance on a wide range of mathematical tasks. More importantly, our analysis reveals the mechanisms behind these improvements. By reinforcing abstract reasoning loops rather than relying on domain-specific techniques, our approach supports the development of general-purpose reasoning capabilities that extend beyond the training puzzles and traditional in-distribution benchmarks.

\section{Related Work}

Reinforcement learning (RL) has demonstrated strong potential in training agents to solve structured reasoning tasks. Prior research can be broadly categorized into two streams: \textit{game-based RL}, which explores emergent strategies in rule-based environments with performance primarily evaluated within the game context; and \textit{math-based RL}, which applies RL to formal domains such as algebra or theorem proving, but often relies on symbolic action spaces or external verifiers. Our work bridges these two areas by introducing logic puzzles as a structured yet generalizable curriculum aimed at fostering transferable reasoning skills for mathematical problem solving.

\subsection{Game-Based Reinforcement Learning}

RL has achieved impressive results in strategic gameplay. AlphaZero~\cite{silver2017mastering} and MuZero~\cite{schrittwieser2020mastering} illustrate how self-play and latent dynamics models can support long-term planning in environments such as Go and Chess. However, performance in these settings is typically measured by win rates or numerical scores, offering limited insight into the transferability of the acquired reasoning to formal problem domains.

Beyond traditional board games, RL agents have also been trained on logic puzzles including Riddle, Sokoban, and Tic-Tac-Toe Progressive Matrices~\cite{ racaniere2017imagination,giadikiaroglou2024puzzle}. These tasks reveal emergent behaviors such as search heuristics and relational inference. Nevertheless, such works seldom investigate whether the reasoning acquired in games can generalize to abstract or academic domains like mathematics. They also tend to stop short of evaluating broader cognitive capabilities.

\subsection{Math-Based Reinforcement Learning}

In contrast, math-oriented RL directly targets symbolic problem solving. Previous studies have used RL to discover expression simplification strategies~\cite{dabelow2025symbolic} or to guide inference within formal proof systems~\cite{wang2025kimina,ren2025deepseek,zhang2025leanabell}. More recent efforts have applied RL to generate intermediate steps in mathematical competition problems, or to solve tasks in symbolic integration and differential equations~\cite{chervonyi2025gold,moshkov2025aimo}. These methods, however, often rely on domain-specific rules or external computational solvers, which limits their general applicability to new settings.

\section{Method}

\subsection{Data Construction}

Each puzzle instance in our dataset is designed to meet three core criteria: (1) logical consistency, (2) a unique solution, and (3) controllable difficulty. Logical consistency prevents unsatisfiable states that might impede learning. Uniqueness ensures that the model receives clear and unambiguous feedback. Difficulty control enables curriculum learning, allowing the model to gradually advance from simpler to more complex reasoning tasks. The seven puzzle types we construct share several essential properties, including structured reasoning constraints, verifiable correctness, and parameterizable complexity:

\paragraph{Sudoku} A 9$\times$9 grid must be completed so that every row, column, and 3$\times$3 subgrid contains the digits from 1 to 9 exactly once. Full solutions are generated through backtracking algorithms, after which digits are removed while maintaining uniqueness via a constraint solver. Difficulty is modulated by the number of clues and the sophistication of solving techniques required.

\paragraph{Nonogram} This puzzle involves filling an $N\times N$ grid based on run-length clues provided for each row and column. These clues reveal a hidden binary image. They are derived from predefined patterns and verified to ensure the uniqueness of the solution. The difficulty level depends on grid dimensions, fill density, and visual complexity of the target pattern.

\paragraph{Cryptarithm} In this puzzle, letters within an arithmetic equation (e.g., \texttt{SEND + MORE = MONEY}) must be mapped to distinct digits to satisfy the numerical constraint. Valid digit-to-letter mappings are first created and then concealed, and uniqueness is enforced through symbolic solvers. Puzzle complexity is governed by the number of unique letters and the carry-over intricacies of the equation.

\paragraph{Magic Square} An $n\times n$ grid is filled with integers from 1 to $n^2$ such that the sums of each row, column, and both diagonals are equal. We construct valid configurations using classical generation methods and remove entries selectively to preserve uniqueness. Difficulty is controlled by the value of $n$ and the sparsity of the remaining clues.

\paragraph{Zebra Puzzle} This puzzle requires assigning attribute values (e.g., nationality, house color) to fixed positions based on natural-language constraints. We use templated clues and logic solvers to guarantee unique solutions. The complexity of each instance is adjusted by varying the number of attributes involved and the logical depth needed to resolve them.

\paragraph{Graph Connectivity} Given a graph generated from the Erdős–Rényi model $G(N, p)$, the task is to determine whether the graph is connected. Each instance is associated with a unique binary label and is represented through textual edge lists. We modulate difficulty by varying the number of nodes ($N$) and edge probability ($p$), thereby spanning sparse, critical, and dense regimes.

\paragraph{Knights and Knaves} In this logic puzzle, each character is either a knight (always tells the truth) or a knave (always lies). Given a series of statements made by the characters about themselves or others (e.g., “A says B is a knave”), the goal is to deduce each person’s true identity. We generate consistent truth assignments and derive corresponding statements that uniquely identify them. Difficulty is controlled by the number of characters and the level of logical nesting or indirection involved in the inference process.

\subsection{Training Procedure}

In our approach, each logic game dataset provides ground-truth validation at multiple stages, including format compliance, intermediate reasoning steps, and final solution correctness. This structure enables the construction of a composite reward function that does not rely on game-specific reward magnitudes. Concretely, for a model trajectory

\[
(s_{0}, a_{0}, s_{1}, \dots, s_{T}, a_{T}, s_{T+1}),
\] 
we define the cumulative reward as
\begin{equation}
R \;=\; \sum_{t=0}^{T} \Bigl[r_{\mathrm{fmt}}(s_{t}, a_{t}) \;+\; r_{\mathrm{int}}(s_{t}, a_{t})\Bigr] \;+\; r_{\mathrm{final}}(s_{T+1})\,,
\end{equation}
where 
\begin{itemize}
    \item $r_{\mathrm{fmt}}(s_{t}, a_{t}) \in [0,1]$ penalizes or rewards adherence to the required output format (e.g., correct tokenization, syntactic structure).
    \item $r_{\mathrm{int}}(s_{t}, a_{t}) \in [0,1]$ evaluates the correctness of each intermediate reasoning step, as validated against the dataset’s annotated proofs or partial solutions.
    \item $r_{\mathrm{final}}(s_{T+1}) \in \{0,1\}$ indicates whether the final answer matches the ground‐truth solution.
\end{itemize}
Because each logic game inherently offers step‐by‐step verifiability, we assign nonzero $r_{\mathrm{int}}$ whenever the model’s partial derivation matches a valid intermediate state. This hierarchical reward structure encourages the model to produce well‐formed, logically coherent reasoning chains rather than merely guessing final answers.

Moreover, our datasets permit fine‐grained control over puzzle difficulty. Let $\mathcal{D}_{d}$ denote the subset of problem instances at difficulty level $d\in\{1,\dots,D\}$. During training, we monitor two validation metrics on $\mathcal{D}_{d}$: the average intermediate‐step accuracy $\mathcal{A}_{\mathrm{int}}^{(d)}$ and final‐answer accuracy $\mathcal{A}_{\mathrm{final}}^{(d)}$. When both metrics exceed predefined thresholds $\tau_{\mathrm{int}}$ and $\tau_{\mathrm{final}}$, respectively, we advance to difficulty level $d+1$. Formally, if
\begin{equation}
\mathcal{A}_{\mathrm{int}}^{(d)} \;\ge\;\tau_{\mathrm{int}}
\quad\text{and}\quad
\mathcal{A}_{\mathrm{final}}^{(d)} \;\ge\;\tau_{\mathrm{final}},
\end{equation}
then we set $d \leftarrow d + 1$. This dynamic curriculum ensures that the model is continually challenged, mitigating overfitting on easier instances and fostering progressive acquisition of complex reasoning skills.

For reinforcement learning, we employ the Generalized Regularized Policy Optimization (GRPO) algorithm. Denoting the policy by $\pi_{\theta}(a_t \mid s_t)$ with parameters $\theta$, GRPO maximizes the expected return plus an entropy‐regularization term to encourage exploration:
\begin{equation}
\mathcal{J}(\theta) \;=\; \mathbb{E}_{\pi_{\theta}}\Bigl[\sum_{t=0}^{T} \gamma^{t}\,R_{t}\Bigr] \;+\; \lambda\,\mathbb{H}\bigl(\pi_{\theta}(\cdot \,\vert\, s_t)\bigr),
\end{equation}
where $R_{t} = r_{\mathrm{fmt}}(s_{t}, a_{t}) + r_{\mathrm{int}}(s_{t}, a_{t})$ at step $t$, $\gamma$ is the discount factor, and $\lambda$ is the entropy coefficient. We update $\theta$ by ascending the gradient $\nabla_{\theta}\mathcal{J}(\theta)$ using policy‐gradient estimators subject to trust‐region constraints, as specified by GRPO.

Since each game’s underlying rules and evaluation criteria differ, we retain separate reward components
\[
\bigl\{r_{\mathrm{fmt}}^{(g)},\,r_{\mathrm{int}}^{(g)},\,r_{\mathrm{final}}^{(g)}\bigr\}
\]
for each game $g$. When constructing a combined dataset 
\[
\mathcal{D}_{\mathrm{all}} = \bigcup_{g=1}^{G} \mathcal{D}^{(g)},
\]
we compute the reward for any trajectory drawn from $\mathcal{D}^{(g)}$ using that game’s specific reward functions. Consequently, the All‐Game RL training objective becomes
\begin{equation}
\begin{aligned}
\mathcal{J}_{\mathrm{all}}(\theta) \;=\;& \sum_{g=1}^{G} \mathbb{E}_{\substack{\pi_{\theta}\\ \mathcal{D}^{(g)}}}
\!\Biggl[\sum_{t=0}^{T^{(g)}} \gamma^{t}\bigl(r_{\mathrm{fmt}}^{(g)}(s_{t},a_{t}) + r_{\mathrm{int}}^{(g)}(s_{t},a_{t})\bigr)\Biggr] \\
& + \sum_{g=1}^{G} \mathbb{E}_{\substack{\pi_{\theta}\\ \mathcal{D}^{(g)}}}\!\bigl[r_{\mathrm{final}}^{(g)}(s_{T^{(g)}+1})\bigr]
\;+\; \lambda\,\mathbb{H}\bigl(\pi_{\theta}\bigr).
\end{aligned}
\end{equation}
By preserving each game’s distinct reward structure, the model learns to generalize across multiple forms of logical deduction, ultimately yielding a policy that captures shared reasoning principles while respecting game‐specific nuances.

\section{Experiments and Analysis}
We evaluate the effectiveness of our “play‐to‐learn” RLVR framework on seven diverse mathematics benchmarks: AIME24, GSM8K, MATH, AMC23, OlympiadBench, Gaokao2024, and Minerva‐MATH~\cite{cobbe2021training,hendrycks2021measuring,he2024olympiadbench,lewkowycz2022solving}. These datasets collectively span elementary arithmetic, middle‐school and high‐school contest problems, and university‐level mathematical reasoning, allowing us to assess both breadth and depth of transfer. In what follows, we first present quantitative results on each benchmark and then we conclude with a detailed analysis of how puzzle‐based RL improves performance across problem categories and difficulty levels.

\subsection{Performance on Game Benchmark}

\begin{figure}[htbp]
    \centering
    \includegraphics[width=0.48\textwidth]{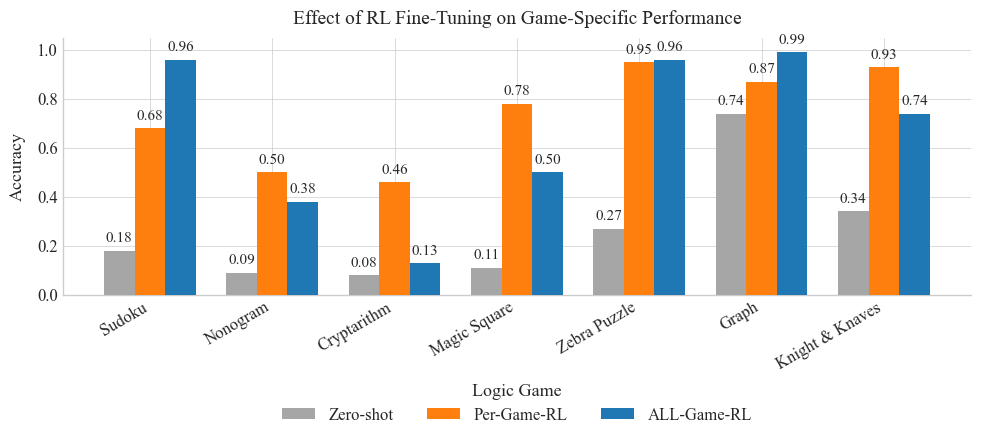} 
    \caption{Performance on Game Benchmarks}
    \label{fig:rl_game_performance}
\end{figure}

Figure~\ref{fig:rl_game_performance} shows consistent performance gains from RL fine-tuning. Individual training yields stronger improvements for Cryptarithm, Magic Square, and KK, where domain-specific heuristics such as symbolic arithmetic or structured layouts play a critical role. In contrast, combined training performs better on Graph, Sudoku, and Zebra, where abstract reasoning strategies like constraint propagation and global consistency tend to generalize across tasks. These trends suggest that individual RL captures task-specialized logic, while multi-task RL fosters reasoning skills that transfer across domains. This highlights their complementary contributions to enhancing mathematical problem-solving. Notably, since these improvements in logic puzzles indicate an increase in fundamental inferential ability, we next investigate how the same RL-enhanced reasoning translates into gains on a range of out-of-domain mathematical benchmarks.

\begin{table*}[]
\centering
\small
\resizebox{\textwidth}{!}{%
\begin{tabular}{lcccccccc}
\toprule
\textbf{Model} & \textbf{AIME24} & \textbf{GSM8K} & \textbf{MATH} & \textbf{AMC23} & \textbf{OlympiadBench} & \textbf{Gaokao2024} & \textbf{Minerva-MATH} & \textbf{Avg. Performance} \\
\midrule
Zero-Shot       & 13.33 & 92.34 & 66.34 & 47.50 & 32.59 & 32.97 & 26.47 & 44.51 \\
Sudoku-RL       & 10.00 & 91.58 & 66.30 & 52.50 & 31.11 & 38.46 & 24.26 & 44.89 (\textcolor{green}{+0.85\%}) \\
Nonogram-RL     & 10.00 & 92.34 & 66.38 & 62.50 & 29.04 & 43.96 & 23.53 & 46.82 (\textcolor{green}{+5.19\%}) \\
Cryptarithm-RL  & 13.33 & 92.87 & 66.82 & 57.50 & 31.11 & 30.77 & 22.43 & 44.98 (\textcolor{green}{+1.06\%}) \\
Magic Square-RL     & 16.67 & 91.96 & 67.26 & 52.50 & 31.26 & 35.16 & 21.32 & 45.16 (\textcolor{green}{+1.46\%}) \\
Zebra-RL        & 10.00 & 91.58 & 66.02 & 52.50 & 32.15 & 31.87 & 23.53 & 43.95 (\textcolor{red}{-1.26\%}) \\
Graph-RL        & 13.33 & 92.49 & 66.56 & 55.00 & 32.59 & 31.87 & 25.00 & 45.26 (\textcolor{green}{+1.68\%}) \\
Knights\&Knaves-RL           & 20.00 & 92.11 & 66.50 & 47.50 & 31.41 & 39.56 & 23.16 & 45.75 (\textcolor{green}{+2.78\%}) \\
All-Game RL     & 20.00 & 91.58 & 67.26 & 55.00 & 31.85 & 47.25 & 24.26 & 48.17 (\textcolor{green}{+8.22\%}) \\
\bottomrule
\end{tabular}
}
\caption{Accuracy (\%) on math benchmarks (columns) of LLMs fine-tuned via RL on different logic games (rows). “Zero-Shot” is the base model before any RL; “All-Game RL” is trained on all games; others are per-game RL. Numbers in parentheses show relative gain over Zero-Shot.}
\label{tab:math_transfer_by_game}
\end{table*}
\subsection{Performance on Math Benchmark}

Table~\ref{tab:math_transfer_by_game} presents zero-shot baseline accuracy (44.51\%) alongside results from models fine-tuned via reinforcement learning (RL) on individual logic games and a combined ``All-Game'' curriculum. Six of the seven game-specific RL models outperform the baseline: \textit{Graph-RL} achieves 45.26\% (+1.68\%), \textit{Cryptarithm-RL} 44.98\% (+1.06\%), \textit{Magic Square-RL} (Magic Square) 45.16\% (+1.46\%), \textit{Sudoku-RL} 44.89\% (+0.85\%), \textit{Knight \& Knaves} 45.75\% (+2.78\%) and \textit{Nonogram-RL} 46.82\% (+5.19\%), with the latter benefiting from enhanced performance on grid-based tasks such as \textsc{AIME24} and \textsc{Gaokao2024}. In contrast, \textit{Zebra-RL} underperforms (43.95\%, $-1.26\%$), suggesting that positional logic heuristics from zebra puzzles do not transfer effectively to the target mathematical domains.

The \textit{All-Game RL} model, trained jointly on all seven tasks, achieves the highest average accuracy of 48.17\% (+3.66\% absolute; +8.22\% relative), demonstrating that a diverse training curriculum fosters more generalizable inference capabilities. For example, accuracy on \textsc{Gaokao2024} improves from 32.97\% (zero-shot) to 47.25\% (+14.28\% absolute), and on \textsc{AIME24} from 13.33\% to 20.00\% (+6.67\% absolute). These improvements reflect enhanced abilities in eliminating infeasible assignments, maintaining global consistency, and executing multi-step deductions.

In sum, RL fine-tuning on logic games strengthens core reasoning skills such as constraint propagation, consistency enforcement, and sequential planning. These capabilities translate directly into improved mathematical problem solving. While individual game-based RL offers modest to notable gains (up to +5.19\%), the \textit{All-Game} curriculum consistently yields superior performance by exposing the model to a broader range of deductive patterns. These findings support the conclusion that augmenting LLMs with RL-based logical training is an effective approach to enhancing their mathematical reasoning abilities.

\subsection{Performance across various categories}
\begin{figure}[htbp]
    \centering
    \includegraphics[width=0.45\textwidth]{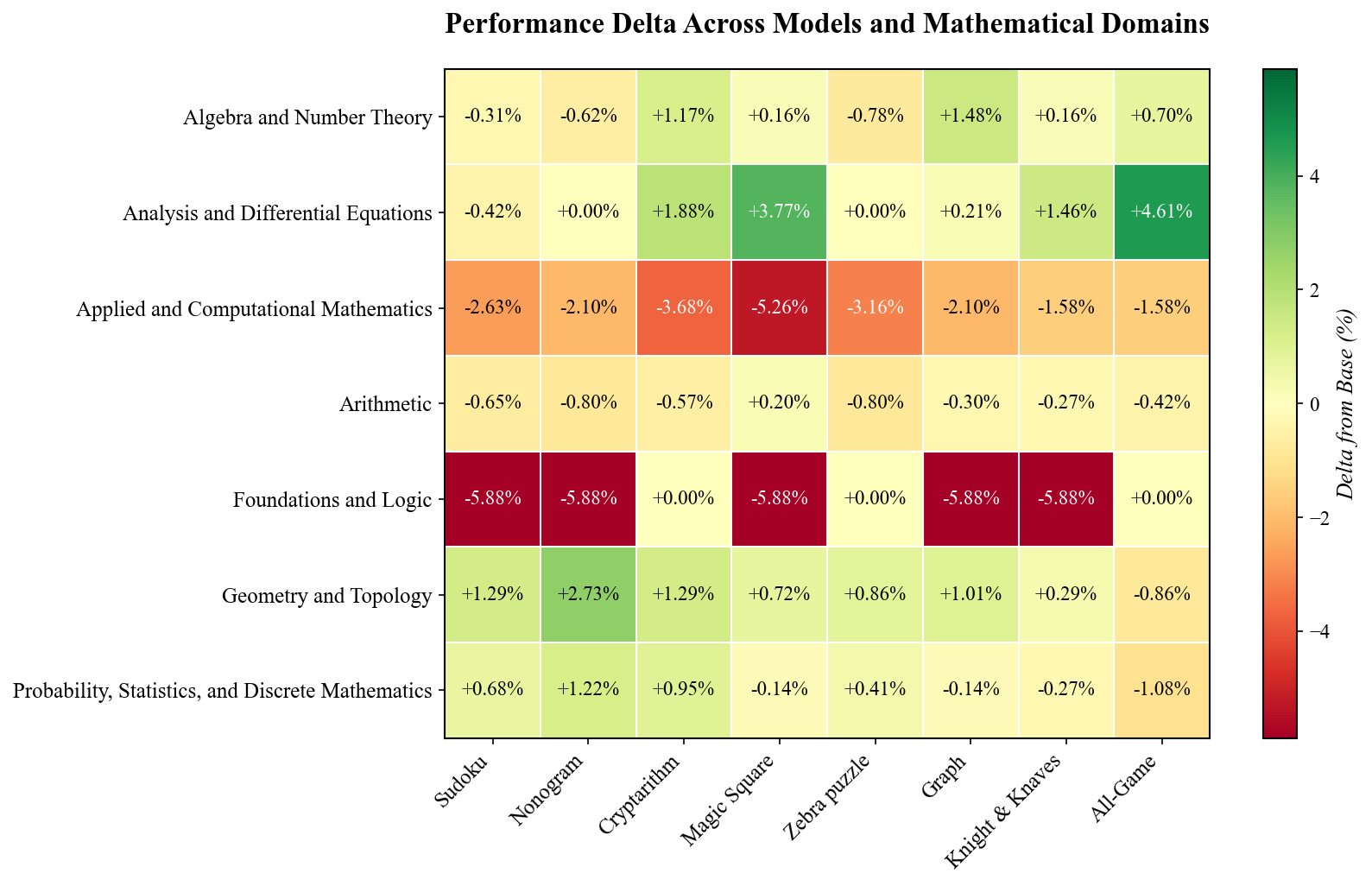} 
    \caption{Performance across various categories}
    \label{fig:category}
\end{figure}

Having established that RL fine tuning on logic puzzles yields strong overall gains on out of domain mathematics benchmarks, we now turn to a more detailed analysis of how these improvements manifest across specific problem categories. As shown in Figure~\ref{fig:category}, structured logic puzzles promote the development of reusable reasoning subroutines that transfer effectively to diverse mathematical tasks.

Puzzles such as Sudoku, Kakuro, Cryptarithm, and Magic Square rely heavily on constraint propagation. Through these tasks, the model learns to eliminate invalid options and gradually refine partial solutions until a unique answer is identified. This encourages a prune and search strategy that involves applying local rules, verifying global consistency, and exploring only viable candidates. As a result, performance improves on algebraic and number theoretic problems. For example, elimination techniques and carry handling heuristics developed through Cryptarithm or Kakuro enhance the model’s ability to solve tasks involving modular arithmetic and symbolic manipulation. Similarly, Magic Square puzzles reinforce multi directional balancing, which resembles the reasoning used in verifying multi term identities or solving differential equations.

Spatial puzzles like Nonogram, and to a lesser extent Sudoku, improve the model’s capacity for two dimensional consistency. In Nonogram, solving requires reconciling clues from rows and columns to reconstruct a coherent image. This spatial reasoning directly supports tasks in geometry and topology, where local changes influence global structure. Notably, training on Nonogram yielded the largest gains in geometric reasoning, suggesting that puzzles requiring simultaneous local and global inference help the model form richer spatial representations.

However, not all puzzles contribute equally across domains. In the Foundations and Logic category, the most significant gains came from puzzles that resemble predicate logic inference. For instance, Zebra puzzles rely on categorical elimination, while Cryptarithms support symbolic assignment learning. In contrast, puzzles involving numeric grids such as Sudoku and Kakuro occasionally hindered performance in logic focused domains, likely because their emphasis on arithmetic patterns shifted the model’s attention away from symbolic inference. Similarly, in Applied and Computational Mathematics, performance declined slightly across all puzzle types, suggesting that purely combinatorial reasoning does not directly support tasks related to algorithmic complexity or numerical stability. These trade offs underscore the importance of aligning puzzle based training with the reasoning requirements of each target category, so that each phase of RL fine tuning reinforces rather than distracts from the intended mathematical abilities.

\subsection{Performance across various difficulties}

\begin{figure}[htbp]
    \centering
    \includegraphics[width=0.45\textwidth]{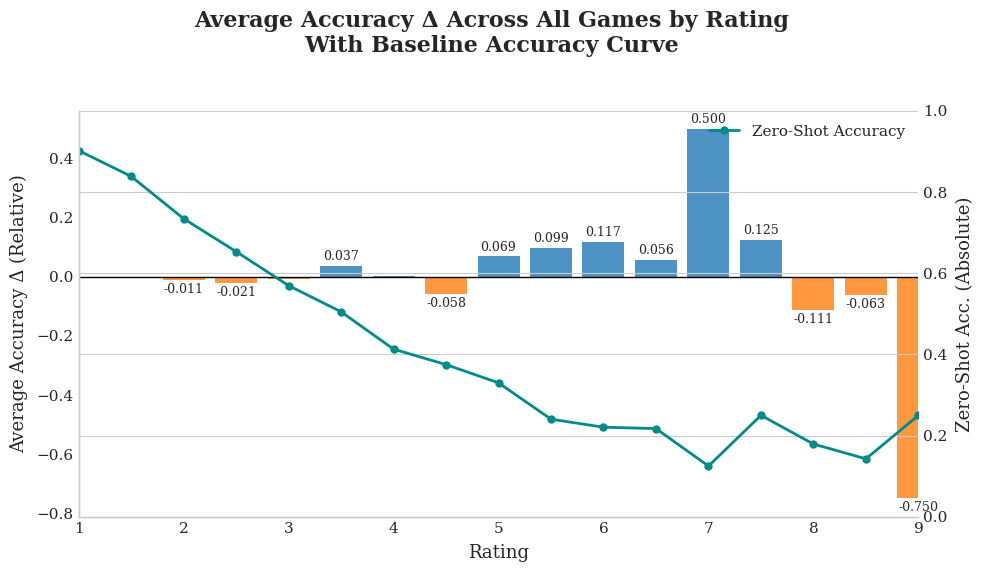} 
    \caption{Performance across various difficulties}
    \label{fig:difficulty}
\end{figure}

Building on the category level analysis, we next investigate how RL fine tuning on logic puzzles affects performance as a function of problem difficulty. Figure~\ref{fig:difficulty} shows that, after RL fine tuning on logic puzzles, the model’s relative accuracy gains peak at mid level difficulty (ratings 5 to 7.5), while performance remains essentially flat or slightly declines on both easy (1 to 3.5) and hard (8 to 9) problems. This pattern supports the interpretation that puzzle based training primarily strengthens general reasoning routines such as multi step deduction and constraint propagation, rather than improving raw calculation skills or instilling rare, domain specific tactics.

Mid difficulty problems typically demand systematic elimination and chained inference, directly aligning with the strategies reinforced by games like Cryptarithm, Sudoku, and Graph Connectivity. As the model internalizes prune and search patterns from these puzzles, it achieves higher accuracy on moderately challenging tasks that require combining local deductions into a global solution.

By contrast, easy problems rely predominantly on straightforward arithmetic or fact recall, where the zero shot model already performs strongly. Puzzle training can even slightly diminish performance in this range by shifting the model’s focus away from memorized shortcuts and toward more elaborate inference procedures. At the highest difficulty levels (ratings 8 to 9), many questions demand niche tricks or domain specific insights, such as nonstandard inequalities or sophisticated geometry constructions, that go beyond the general reasoning routines learned from logic puzzles. Moreover, the relative scarcity of such high rated examples yields greater variance in empirical accuracy, making any gains less reliable.

In summary, logic puzzle RL fine tuning enhances the model’s compositional reasoning and yields the greatest benefit on problems that reward structured deduction. However, it provides limited value for tasks dominated by rote computation or those requiring highly specialized strategies at the extremes of difficulty.

\section{Conclusion}

In this paper, we showed that fine tuning an LLM with reinforcement learning on a suite of logic puzzle games not only yields near perfect in domain performance, but also cultivates general reasoning subroutines—constraint propagation, global consistency checks, and multi step deduction—that transfer to out of domain mathematics benchmarks. Models trained on individual puzzles improved moderately in targeted categories (e.g., Cryptarithm RL bolstered Algebra and Number Theory, Nonogram RL boosted Geometry and Topology), while a joint (All Game) curriculum produced the largest overall uplift (+8.22\% average gain) by exposing the LLM to a broad spectrum of deductive patterns.

Moreover, we found that these game derived reasoning skills yield the greatest benefit on mid level to moderately hard math problems (ratings 5 to 7.5), which demand systematic elimination and chained inference. In contrast, gains on easy arithmetic tasks (ratings 1 to 3.5) and on very difficult, contest style problems (ratings 8 to 9) were negligible or slightly negative, since the former rely on rote calculation and the latter on niche, domain specific tricks. Altogether, our experiments confirm that RL over logic puzzles effectively enhances an LLM’s compositional reasoning, resulting in measurable improvements on diverse mathematical tasks without requiring task specific symbolic frameworks.

\bibliography{aaai2026}

\end{document}